\documentclass[nohyperref]{article}

\usepackage{microtype}
\usepackage{graphicx}
\usepackage{subfigure}
\usepackage{booktabs} 

\usepackage{hyperref}

\usepackage[accepted]{src/icml2022}

\usepackage{amsmath}
\usepackage{amssymb}
\usepackage{mathtools}
\usepackage{amsthm}
\usepackage{caption}

\usepackage[capitalize,noabbrev]{cleveref}

\theoremstyle{plain}

\theoremstyle{definition}

\theoremstyle{remark}

\usepackage[textsize=tiny]{todonotes}

\icmltitlerunning{Simple Regularisation for Uncertainty-Aware Knowledge Distillation}

\begin{document}

\twocolumn[
\icmltitle{Simple Regularisation for Uncertainty-Aware Knowledge Distillation}




\begin{icmlauthorlist}
\icmlauthor{Martin Ferianc}{yyy}
\icmlauthor{Miguel Rodrigues}{yyy}

\end{icmlauthorlist}

\icmlaffiliation{yyy}{Department of Electronic and Electrical Engineering, University College London, London, UK}

\icmlcorrespondingauthor{Martin Ferianc}{martin.ferianc.19@ucl.ac.uk}

\icmlkeywords{}

\vskip 0.3in
]



\printAffiliationsAndNotice{The code is at: \url{https://github.com/martinferianc/hydra_plus}. This is a work-in-progress.} 

\begin{abstract}
Considering uncertainty estimation of modern neural networks (NNs) is one of the most important steps towards deploying machine learning systems to meaningful real-world applications such as in medicine, finance or autonomous systems. 
At the moment, ensembles of different NNs constitute the state-of-the-art in both accuracy and uncertainty estimation in different tasks. 
However, ensembles of NNs are unpractical under real-world constraints, since their computation and memory consumption scale linearly with the size of the ensemble, which increase their latency and deployment cost. 
In this work, we examine a simple regularisation approach for distribution-free knowledge distillation of ensemble of machine learning models into a single NN.
The aim of the regularisation is to preserve the diversity, accuracy and uncertainty estimation characteristics of the original ensemble without any intricacies, such as fine-tuning.  
We demonstrate the generality of the approach on combinations of toy data, SVHN/CIFAR-10, simple to complex NN architectures and different tasks.  
\end{abstract}

\section{Introduction}\label{sec:introduction}

Neural networks (NNs) have enjoyed overwhelming interests in the recent past, due to their automatic feature learning abilities translating to super-human accuracy~\cite{lecun2015deep}. 
However, the deployment of the NNs in the real-world requires more than high accuracy. 
The NN-based system needs to be trustworthy, especially when presented with data that it has previously not observed.
Trust can be built through estimating the uncertainty of the model and its estimation enables the users to gauge whether the model is wrong or lacks sufficient knowledge to solve the task at hand~\cite{bhatt2021uncertainty}. 
\begin{figure}[t!]
    \centering
    \includegraphics[width=1\linewidth]{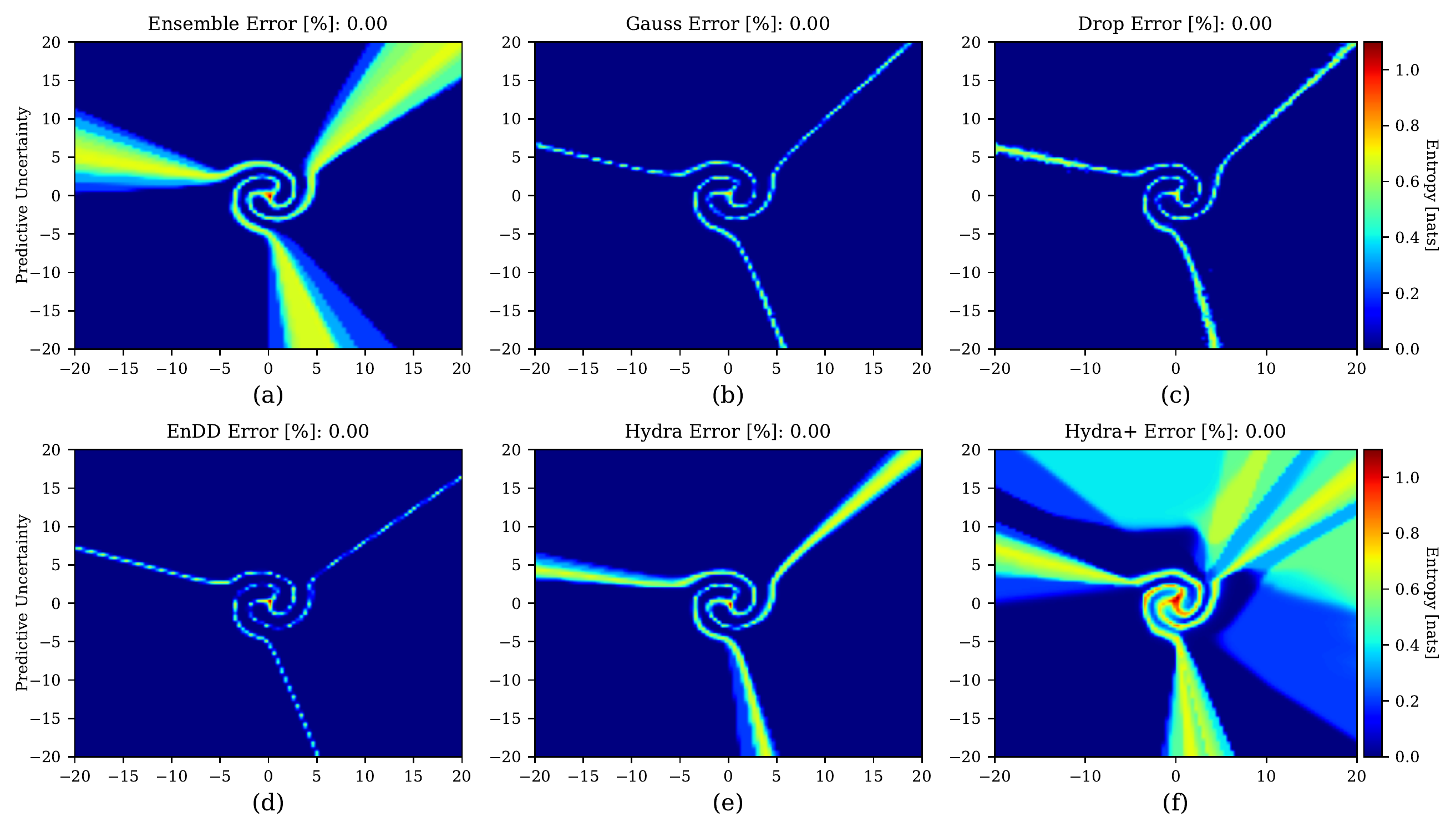}
    \includegraphics[width=0.9\linewidth]{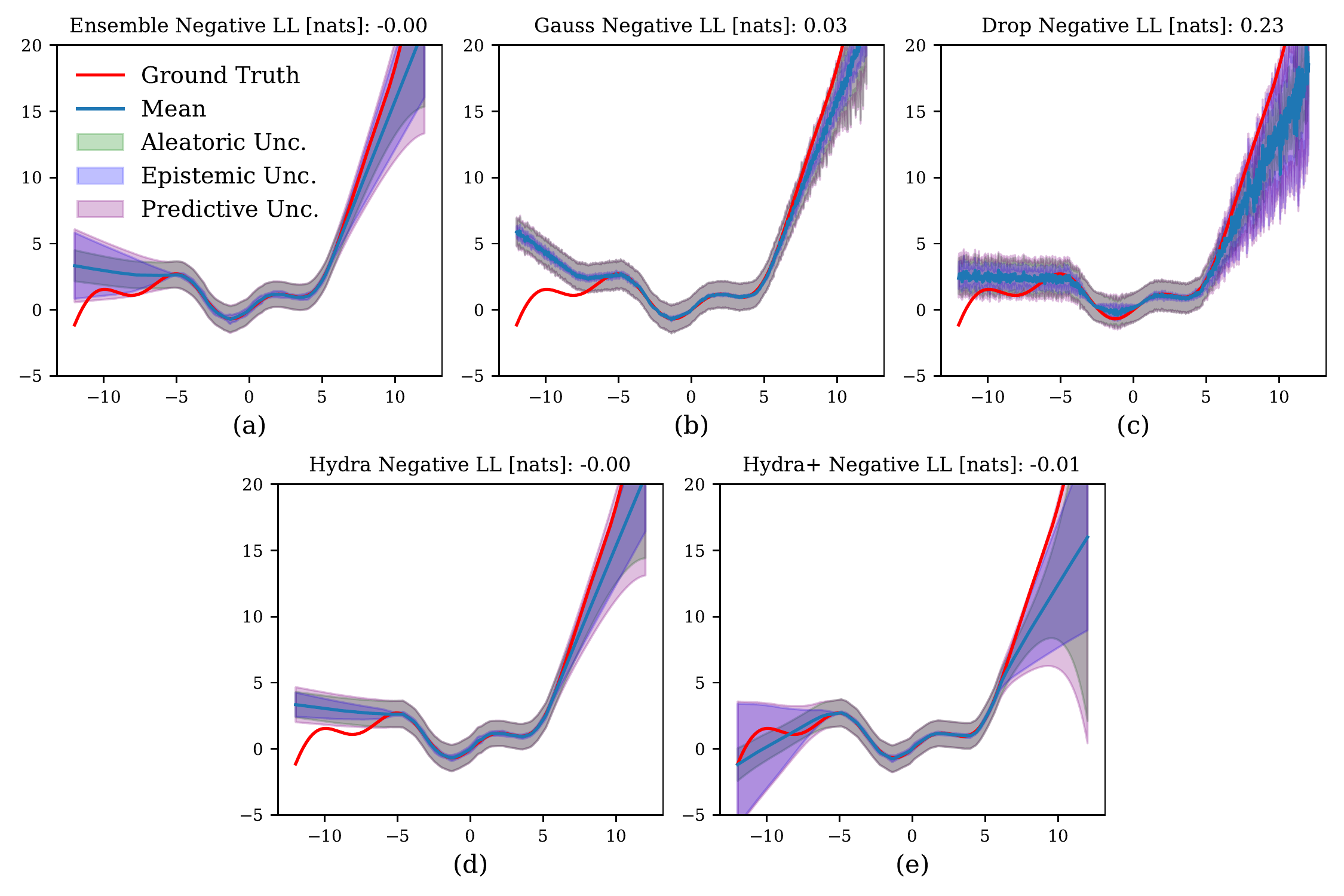}
    \vspace{-0.5em}
    \caption{Toy classification/regression demonstrating KD of an \textit{Ensemble} with $N=20$ members into a student, while comparing different methods: \textit{Gauss},  \textit{Drop}, \textit{EnDD}, \textit{Hydra} and proposed \textit{Hydra+} (a-f, a-e). The plots compare predictive, epistemic or aleatoric uncertainties, with one standard deviation for regression. The titles show the error/negative log-likelihood (LL) on test samples/curve. See Sections~\ref{sec:related_work} or \ref{sec:experiments} for details. The decomposition of aleatoric and epistemic uncertainty in classification is in the Appendix~\ref{sec:appendix}.}
    \label{fig:toy_example}
    
\end{figure}
Therefore, considering uncertainty estimation in modern NNs is increasingly important especially for safety-critical applications such as in healthcare or self-driving~\cite{abdar2021review}.

At the moment, ensembles of NNs~\cite{zaidi2021neural} provide the best quantitative and qualitative results in terms of accuracy or uncertainty estimation. 
The ensemble can be created as simply as training different machine learning models with different seeds, meaning different initialisation of their parameters. 
The initialisation differentiation facilitates distinct optimisation trajectories of the ensemble members, ending in diverse local minima, which gives ensembles their representation capacity~\cite{lakshminarayanan2016simple}. 
Then, it is possible to estimate the uncertainty of the complete ensemble through the disagreement of its individual members. 
In comparison to rigorous distribution-free uncertainty estimation methods~\cite{angelopoulos2021gentle}, the user only needs to have access to the training data and train a machine learning model several times without any other assumptions to form an ensemble.

However, deploying ensembles of NNs in the real-world constitutes a challenge, since their resource consumption scales linearly with the size of the ensemble. 
\textit{Knowledge distillation} (KD)~\cite{hinton2015distilling, wang2021knowledge} has been previously successfully utilised to compress the representation of the ensemble to a single NN, without any assumptions about the distilled model.  
Nonetheless, capturing the uncertainty of the ensemble without additional data or fine-tuning remains a challenge even with modern KD methods for capturing the ensemble's uncertainty, shown in Figure~\ref{fig:toy_example}. 

In this work, we build on the \textit{Hydra} KD idea proposed by~\cite{tran2020hydra} and we examine a simple regularisation to improve KD in capturing the uncertainty of the original ensemble, which we denote \textit{Hydra+}.
The regularisation is composed of two parts: \textit{a)} modification of the loss function to capture \textit{correctness}, \textit{aggregated} and \textit{individual} performance of the ensemble; \textit{b)} minimisation of similarity between weights of different predictive heads of the multi-head student NN to promote \textit{diversity}. 
These changes result in improved uncertainty estimation and calibration without requiring additional data, modelling assumptions or fine-tuning.
We demonstrate the generality of the examined approach with respect to classification and regression on toy-data, SVHN, CIFAR-10 and simple feed-forward or convolutional, residual~\cite{he2016deep} architectures.

\section{Preliminaries \& Related Work}\label{sec:related_work}

We now cover the preliminaries about ensembles, knowledge distillation and uncertainty decomposition. We also overview the related work.

\subsection{Preliminaries}

\paragraph{Ensembles} Uncertainty estimation is increasingly gaining traction in the machine learning community in order to boost interpretability of the NNs deployed in the real-world~\cite{bhatt2021uncertainty}. 
Despite different sophisticated attempts~\cite{ovadia2019can}, ensembles maintain the state-of-the-art in uncertainty estimation, without requiring any particular assumptions about the data or the task~\cite{lakshminarayanan2016simple,wenzel2020hyperparameter, zaidi2021neural}. 
The process of building a baseline ensemble is simple - train a set of NNs on the same data, with the same architecture, but initialised with different random seeds~\cite{lakshminarayanan2016simple}. 
Despite their generalisation and uncertainty estimation performance, ensembles are difficult to deploy in practice, since their compute and memory demands scale with complexity $\mathcal{O}(N)$ where $N$ is the size of the ensemble. 

\paragraph{Knowledge Distillation (KD)} 
The deployment challenge of ensembles served as one of the inspirations for the idea of \textit{Knowledge distillation} (KD)~\cite{hinton2015distilling}, which aims to compress a large model, or a set of models to a single, smaller less demanding model. 
In general, KD achieves this through guiding the small model - the student to mimic the behaviour of the large ensemble - the teacher~\cite{wang2021knowledge}. Concretely, the guidance between teacher and student is implemented through minimising the Kullback-Leibler (KL)~\cite{kullback1951information} divergence between the likelihoods of the teacher and the student as: $KL(p(\mathbf{\hat{y}}|\mathbf{x},\mathbf{\theta}_T)\|p(\mathbf{\hat{y}}|\mathbf{x},\mathbf{\theta}_S))$ where $\mathbf{\hat{y}}, \mathbf{x}$ are the output prediction and input and $\mathbf{\theta}_T$, $\mathbf{\theta}_S$ are the parametrisations of the ensemble and the student respectively. In finer granularity, the guidance can be pointed towards different characteristics of the teacher, which is the investigation of this work. Namely, our focus is on the ability of the student to capture both the generalisation performance and aleatoric and epistemic uncertainty~\cite{hullermeier2021aleatoric} of the ensemble as close as possible.

\paragraph{Uncertainty Decomposition} 
Unless the knowledge of the data generating process is known, the proposed models always contain a notion of uncertainty~\cite{hullermeier2021aleatoric}. Furthermore, this uncertainty can be decomposed into uncertainty relating to incorrect model assumptions: \textit{epistemic} or noisy data: \textit{aleatoric} uncertainty. The uncertainty's decomposition enables practitioners to understand what the model does not know.
For example, in classification, if $D$ is the training dataset, $\mathbf{\hat{y}}$ are the softmax probabilities produced by a model on data $\mathbf{x}$, parametrised by $\mathbf{\theta}$ as $p(\mathbf{\hat{y}}|\mathbf{x},\mathbb{\theta})$ and $\mathbb{H}, \mathbb{E}$ and $\mathbb{MI}$ are the entropy, expectation and mutual information operators, it corresponds to~\cite{hullermeier2021aleatoric}:

\vspace{-2em}

\begin{multline}
    \underbrace{\mathbb{H}(\mathbb{E}_{p(\mathbb{\theta}|D)}[p(\mathbf{\hat{y}}|\mathbf{x},\mathbb{\theta})])}_\text{Predictive Unc.} =\underbrace{\mathbb{E}_{p(\mathbb{\theta}|D)}[\mathbb{H}(p(\mathbf{\hat{y}}|\mathbf{x},\mathbb{\theta})]}_\text{Aleatoric Unc.} 
    \\+ \underbrace{\mathbb{MI}[\mathbf{\hat{y}},\mathbf{\theta}|\mathbf{x},D]}_\text{Epistemic Unc.}
    \label{eq:unc_decomposition_class}
\end{multline}

\vspace{-0.5em}

In practice the $\mathbb{MI}$ term cannot be computed in a closed-form in NNs, but it can be simply computed by subtracting the aleatoric uncertainty from the predictive uncertainty. Additionally, the expectations are often approximated with empirical averages using $N$ Monte Carlo samples from the learnt posterior distribution $p(\mathbf{\theta}|D)$ for that given method. In ensembles, the samples would correspond to the $N$ different learnt models.

If considering regression and Gaussian likelihood, then:

\vspace{-1em}

\begin{equation}
     \underbrace{\mathbf{\sigma}^2}_\text{Predictive Unc.} = \underbrace{\mathbf{\hat{\sigma}}^2}_\text{Aleatoric Unc.} + \underbrace{\mathbb{V}_{p(\mathbb{\theta}|D)}[p(\mathbf{\hat{y}}|\mathbf{x},\mathbb{\theta})]}_\text{Epistemic Unc.}
     \label{eq:unc_decomposition_reg}
\end{equation} 

\vspace{-0.5em}

$\mathbf{\hat{\sigma}^2}$ is the predicted aleatoric variance, $\mathbf{\hat{y}}$ is the predicted mean and $\mathbb{V}$ is the variance operator. 
In this work, we concentrate specifically on the KD methods that focus on distilling aleatoric and epistemic uncertainty, along with generalisation performance, of the teacher, irrespective of the teacher training procedure or its architecture.

\subsection{Related Work}

In distillation of uncertainty,~\citet{malinin2019ensemble} propose \textit{EnDD} by using a prior network~\cite{malinin2018predictive} as the student, however, their approach requires further fine-tuning on auxiliary data to fully capture the ensemble's uncertainty and it works only for classification problems. \citet{tran2020hydra} proposed \textit{Hydra}: a multi-headed model where each head is paired with a member of the ensemble, while reusing a shared core architecture. The heads aim to capture the diversity of the ensemble, while reusing the common features. However, their approach requires multiple-steps, fine-tuning and inflexibility in choosing the student architecture. \citet{lindqvist2020general, shen2021real} propose a simple distillation method for learning the conditional predictive distribution of the ensemble or a Bayesian NN, into a flexible parametric distribution modelled by the last layer of the NN. For comparison, we consider two such distributions, through adding dropout before the last layer~\cite{gal2016dropout} or using the local-reparametrisation trick with a Gaussian mean-field prior~\cite{kingma2015variational} and we denote them as \textit{Drop} and \textit{Gauss} respectively. The downside of parametrising a distribution through the last layer is that it is necessary to assume some prior distribution, which can be unintentionally misspecified~\cite{fortuin2022priors}. In summary, ensemble constitutes the baseline assumption-free performance which distribution-specified \textit{EnDD}, \textit{Drop} and \textit{Gauss} and assumption-free: \textit{Hydra} or \textit{Hydra+} KD methods target to match. \textit{Hydra+} builds on~\cite{tran2020hydra} and attempts to avoid multi-step training while maintaining the generalisation and improving the uncertainty estimation performance through modifying the training loss function and including a diversity-inducing term.

\section{Method}\label{sec:method}

We now define the loss decomposition along with the diversity-inducing regularisation applied during training.

\subsection{Loss Decomposition}\label{sec:method_loss_decomposition}

The goal of KD is to match the performance of the teacher ensemble with $N$ members, parametrised by $\mathbf{\theta}_{T} =\{\mathbf{\theta}^n_{T}\}_{n=1}^N$ through a student parametrised by $\mathbf{\theta}_S$ on data tuples $(\mathbf{x}, \mathbf{y})$ initially coming from training set $D$, where $\mathbf{x}$ and $\mathbf{y}$ are the input and the desired output. If the end task is classification, e.g. categorisation of images into some classes, the output are the probabilities of the one-hot encoded labels $\mathbf{y}\in\mathbb{R}^{K}$ with $K$ classes as $\textrm{Cat}(\mathbf{y}|\mathbf{\hat{y}}_{T})$ with $\mathbf{\hat{y}}_T\in\mathbb{R}^{N\times K}$ or $\textrm{Cat}(\mathbf{y}|\mathbf{\hat{y}}_S)$ and $\mathbf{\hat{y}}_S\in\mathbb{R}^{K}$ for the teacher and the student respectively, provided that $\mathbf{\hat{y}}$ are obtained through the softmax activation at the output of both ensemble members and the student. If the task is regression, e.g. a prediction of a stock price, the output in both instances $y\in \mathbb{R}$ is modelled as a Gaussian with mean and aleatoric variance as $\mathcal{N}(y|\mathbf{\hat{\mu}}_{T}, \mathbf{\hat{\sigma}}^{2}_T)$ with $\mathbf{\hat{\mu}}_T,\mathbf{\hat{\sigma}}^2_T\in\mathbb{R}^{N}$ or $\mathcal{N}(y|\mathbf{\hat{\mu}}_S, \mathbf{\hat{\sigma}}^2_S)$ with $\mathbf{\hat{\mu}}_S,\mathbf{\hat{\sigma}}^2_S\in\mathbb{R}$ for the teacher and the student.

\begin{figure}[t]
    \centering
    \includegraphics[width=1.0\linewidth]{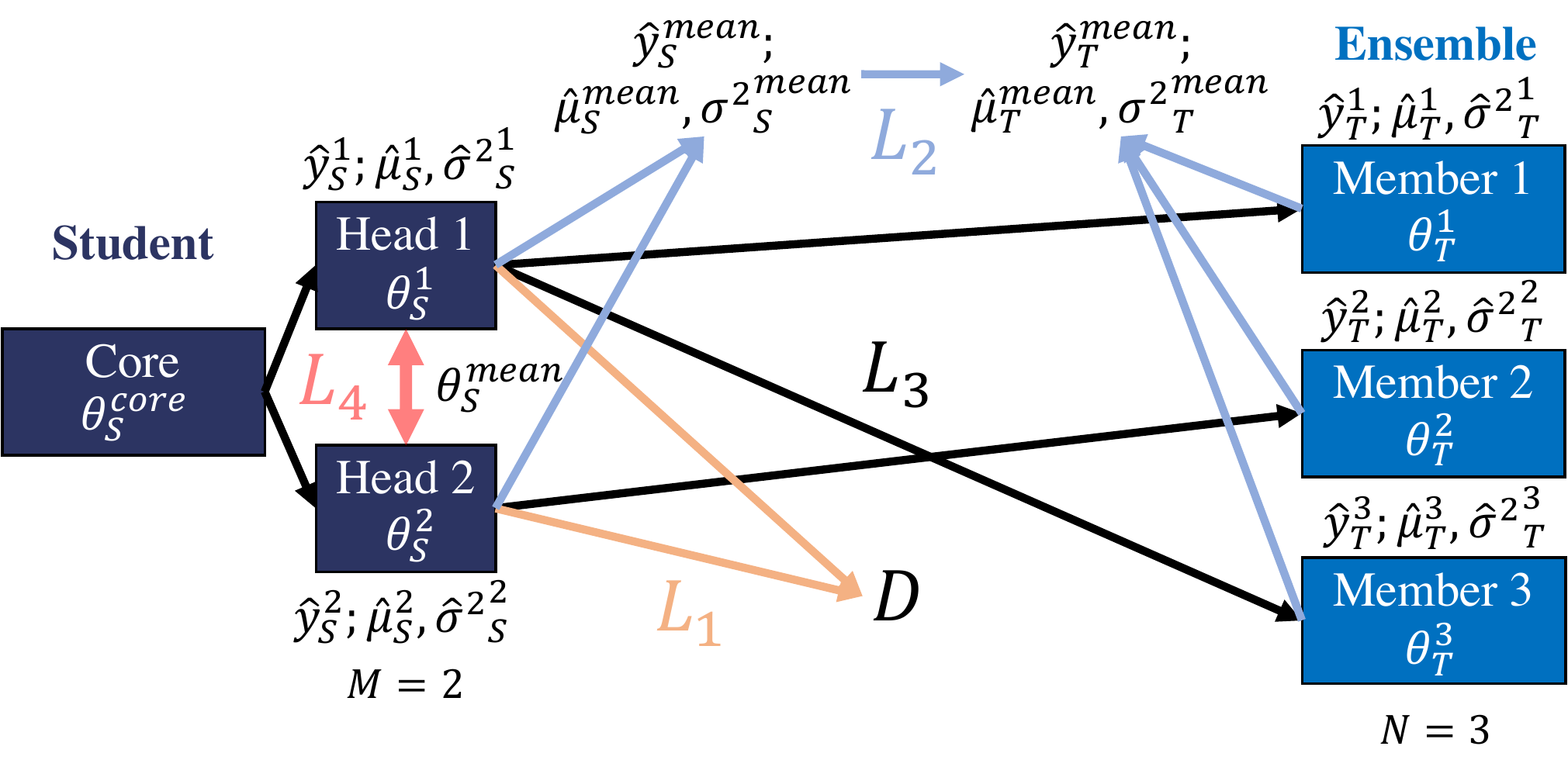}
    \vspace{-2.2em}
    \caption{Student with a core and $M=2$ heads is matched with ensemble consisting of $N=3$ members. Training is performed with respect to the 4 loss components: $L_1$ correctness,  $L_2$ aggregation, $L_3$ individuality and $L_4$ diversity.}
    \label{fig:method}
\end{figure}

Inspired by \textit{Hydra}~\cite{tran2020hydra}, in this work we decompose the student network into two parts, the shared core parametrised by $\mathbf{\theta}_S^{\textrm{core}}$ and $M$ heads with the same structure such that $\mathbf{\theta}_S=\{\mathbf{\theta}_S^{\textrm{core}};\{\mathbf{\theta}^m_{S}\}_{m=1}^{M}\}$. 
The purpose of the core part of the network is to capture common features, while the $M$ heads are supposed to capture the individual intricacies of the teacher ensemble members and enable the decomposition of the prediction into aleatoric and epistemic uncertainty, since the output is $\mathbf{\hat{y}}_S\in\mathbb{R}^{M\times K}$ and $\mathbf{\hat{\mu}}_S,\mathbf{\hat{\sigma}}^2_S\in\mathbb{R}^{M}$. 
The relationship between the teacher and the student along with the used notation is visualised in Figure~\ref{fig:method}. 
Next, we introduce the 4 components $L_1, L_2, L_3, L_4$ of the proposed loss function $L$ that aims to promote \textit{correctness} in the student and capture \textit{aggregated} and \textit{individual} behaviour of the ensemble together with its \textit{diversity}.

\paragraph{Correctness} While proposing KD~\citet{hinton2015distilling} noted that it is necessary to ensure correctness of the student by not only teaching it to mimic the behaviour of the teacher but also making it correct with respect to the training labels or values depending on classification or regression. We adopt this notion with respect to each student head's $m$ output $\hat{y}_S^m; \hat{\mu}_S^m, \hat{\sigma}{^2}^m_S$, where each head should be independently correct. This concept was previously considered in parameter-shared ensembles~\cite{lee2015m}. For classification with the correct label $y_k$ for class $k$ it corresponds to minimising the mean of the cross-entropy across the $M$ heads:

\vspace{-2.em}

\begin{equation}
    L_1^{class}(\mathbf{\hat{y}}_S; y_k) = -\frac{1}{M}\sum_{m=1}^My_k\log{\hat{y}_S^m}
\label{eq:l1_class}
\end{equation}

\vspace{-0.5em}

For regression for target $y$ and unit variance the Gaussian negative log-likelihood reduces to:

\vspace{-2em}

\begin{multline}
    L_1^{reg}(\hat{\mu}_S^m, \hat{\sigma}{^2}^m_S; y) =\\
    \frac{1}{2M}\sum_{m=1}^M \left(  \frac{(\hat{\mu}_S^m - y)^2}{\hat{\sigma}{^2}^m_S} +\log\hat{\sigma}{^2}^m_S \right) 
    + \text{C}
\label{eq:l1_reg}
\end{multline}

\vspace{-0.5em}

C is a constant term not affecting $\mathbf{\theta}_S$ during optimisation.

\paragraph{Aggregation} Next,~\citet{hinton2015distilling} proposed the distillation itself where the output of the ensemble is averaged with respect to $N$ to capture its overall behaviour. For classification this means $\mathbf{\hat{y}}^{mean}_T=\frac{1}{N}\sum_{n=1}^{N}\mathbf{\hat{y}}_T^n$. Similarly, we average the student output $\mathbf{\hat{y}}^{mean}_S=\frac{1}{M}\sum_{m=1}^{M}\mathbf{\hat{y}}_S^m$ where the loss is then the minimisation of the rearranged KL divergence between teacher and student outputs as:

\vspace{-2em}

\begin{multline}
    L_2^{class}(\mathbf{\hat{y}}^{mean}_T; \mathbf{\hat{y}}^{mean}_S) = \\ -\sum_{k=1}^K \hat{y}^{k, mean}_T \log\hat{y}^{k, mean}_S + C
    \label{eq:l2_class}
\end{multline}

\vspace{-1.em}

the logits of the student and teacher can also be softened via division with temperature $T_{mean}\geq1$ prior to softmax to give $\mathbf{\hat{y}}_T$ or $\mathbf{\hat{y}}_S$. 

For regression, we decide to aggregate the teacher and student prediction as means $\hat{\mu}^{mean}_T=\frac{1}{N}\sum_{n=1}^{N}\mathbf{\hat{\mu}}_T^n$ and $\hat{\mu}^{mean}_S=\frac{1}{M}\sum_{m=1}^{M}\mathbf{\hat{\mu}}_S^m$. However, for variance we not only take the mean of the aleatoric variance but we calculate the variance of both the teacher and student prediction to give $\sigma{^2}^{mean}_T =\frac{1}{N}\sum_{n=1}^{N}\hat{\sigma}{^2}^n_T + \mathbb{V}[\mathbf{\hat{\mu}}_T]$ and $\sigma{^2}^{mean}_S =\frac{1}{M}\sum_{m=1}^{M}\hat{\sigma}{^2}^m_S + \mathbb{V}[\mathbf{\hat{\mu}}_S]$ and thus we capture the complete predictive uncertainty. Then, the student output is compared to the teacher again via KL divergence between Gaussians: 

\vspace{-2.5em}

\begin{multline}
    L_2^{reg}(\hat{\mu}^{mean}_T, \sigma{^2}^{mean}_T;\hat{\mu}^{mean}_S, \sigma{^2}^{mean}_S) =\\\frac{1}{2}\left( \frac{\sigma{^2}^{mean}_T+ (\hat{\mu}^{mean}_T- \hat{\mu}^{mean}_S)^2}{\sigma{^2}^{mean}_S} + \log\sigma{^2}^{mean}_S\right)+\textrm{C}
    \label{eq:l2_reg}
\end{multline}

\vspace{-2em}

\paragraph{Individuality} The primary loss with respect to which~\citet{tran2020hydra} were training the \textit{Hydra} in their second phase was matching the $M$ individual heads to the $N$ ensemble members, also conditioning that $M=N$. The motivation behind this loss was to urge each head to learn the representation of the individual ensemble member. If only $L_3$ is being used, all observable individuality is lost~\cite{lee2015m}. We relax the equality constraint on the number of heads $M$ in order to explore algorithmic-hardware trade-offs from reducing $M$ such that $2\leq M\leq N$. If $N> M$, the remaining $N-M$ ensemble members are fairly divided between the $M$ heads. Again for classification and outputs of the teacher ensemble $\mathbf{\hat{y}}_T$ and the student $\mathbf{\hat{y}}_S$ this KL divergence between the teacher and the student rearranges to: 

\vspace{-2em}

\begin{multline}
    L_3^{class}(\mathbf{\hat{y}}_T;  \mathbf{\hat{y}}_S) = 
    -\frac{1}{N}\sum_{n=1}^N\sum_{k=1}^K \hat{y}^{k, n}_T \log\hat{y}^{k, n\%M}_S +C
    \label{eq:l3_class}
\end{multline}

\vspace{-1em}

where \% represents the modulo operator. The logits of the student and teacher can again be softened via division with temperature $T_{ind}\geq1$ prior to softmax to give $\mathbf{\hat{y}}_T$ or $\mathbf{\hat{y}}_S$. 

Likewise for regression and outputs $\hat{\mu}_T, \hat{\sigma}^2_T;\hat{\mu}_S, \hat{\sigma}^2_S$ for teacher and student the KL divergence between the teacher and the student can be formulated as:

\vspace{-2em}

\begin{multline}
    L_3^{reg}(\mathbf{\hat{\mu}}_T, \mathbf{\hat{\sigma}}^2_T;\mathbf{\hat{\mu}}_S, \mathbf{\hat{\sigma}}^2_S) =\frac{1}{N}\sum_{n=1}^N\\\frac{1}{2}\left( \frac{\hat{\sigma}{^2}^n_T+ (\hat{\mu}^{n}_T- \hat{\mu}^{n\%M}_S)^2}{ \hat{\sigma}{^2}^{n\%M}_S} + \log\hat{\sigma}{^2}^{n\%M}_S\right)+\textrm{C}
    \label{eq:l3_reg}
\end{multline}

\vspace{-1.em}

\paragraph{Diversity} We empirically observed that it is not possible to induce diversity in one-shot training in the student by using $L_3$ alone. Therefore, we examine a differentiable diversity-inducing term calculated between the weights of the heads of the student $\{\mathbf{\theta}^m_{S}\}_{m=1}^{M}$ at each layer-level $l=1,\dots,L$, where $l=1$ is the first weight-containing layer in the head and $l=L$ is the output layer. 
The core idea is to reduce the similarity between the weights and repulse them from each other at each level in order to obtain diverse responses to the same input processed through the shared core. 

We define the mean head weight at an arbitrary level simply as $\mathbf{\theta}^{mean}_{S}=\frac{1}{M}\sum_{m=1}^{M}\mathbf{\theta}^m_{S}$. Given the abstract mean weight representation, we propose to minimise the similarity between the mean head weight and the individual head weights for each layer-level $l=1,\dots,L$ of the head as:

\vspace{-2em}

\begin{equation}
    L_4(\mathbf{\theta}^{mean}_{S};\{\mathbf{\theta}^m_{S}\}_{m=1}^{M}) = \sum_{l=1}^{L}\sum_{m=1}^{M} \frac{1+\textrm{cos}(\mathbf{\theta}^{l,mean}_{S}, \mathbf{\theta}^{l,m}_{S})}{2}
    \label{eq:l4}
\end{equation}

\begin{figure*}[t]
    \centering
    \includegraphics[width=1\linewidth]{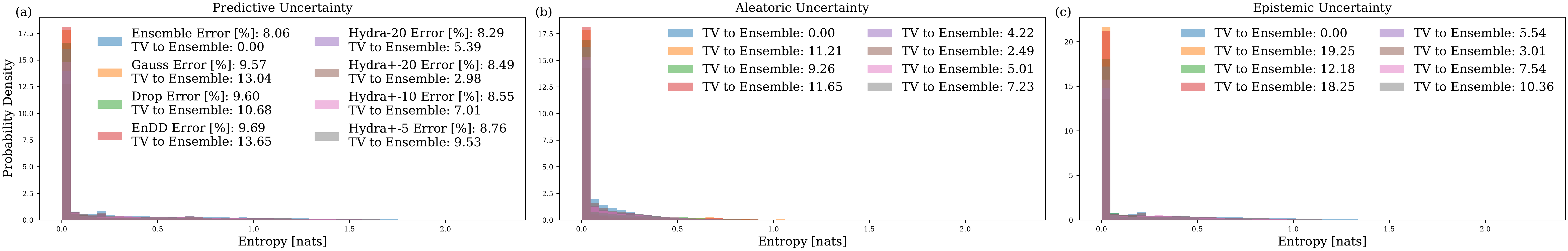}
    \includegraphics[width=1\linewidth]{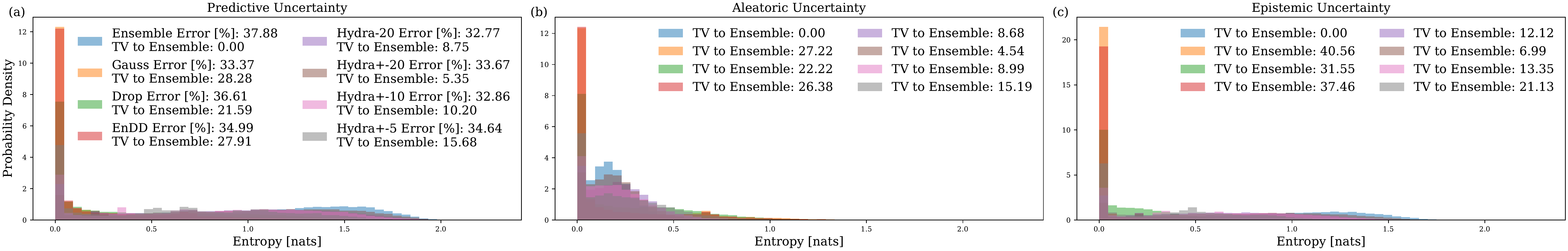}
    \includegraphics[width=1\linewidth]{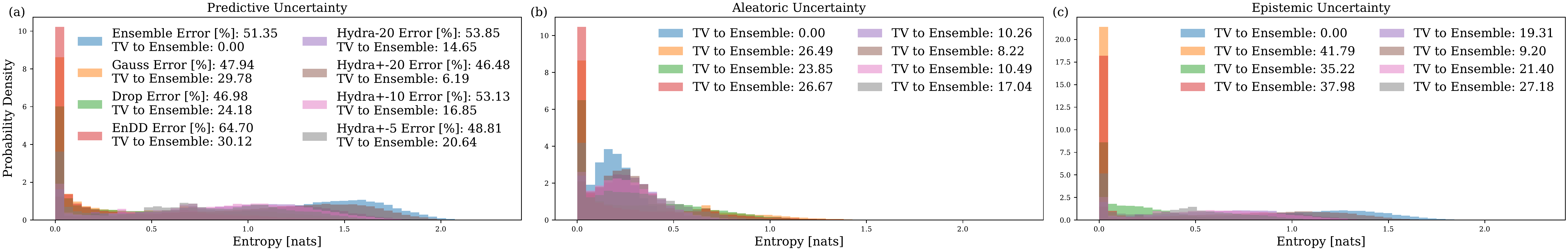}
    \vspace{-2em}
    \caption{Uncertainty decomposition for augmentations applied to the SVHN test set and the compared methods for one seed/experiment. From the top, in the first row (a-c) is increasing the brightness by 30\%, second row (a-c) is the 15° rotation and the third row (a-c) is 20\% vertical shift. TV denotes total variation and Error denotes the error on the augmented test set.}
    \label{fig:svhn}
\end{figure*}

\begin{figure*}[t]
    \centering
    \includegraphics[width=1\linewidth]{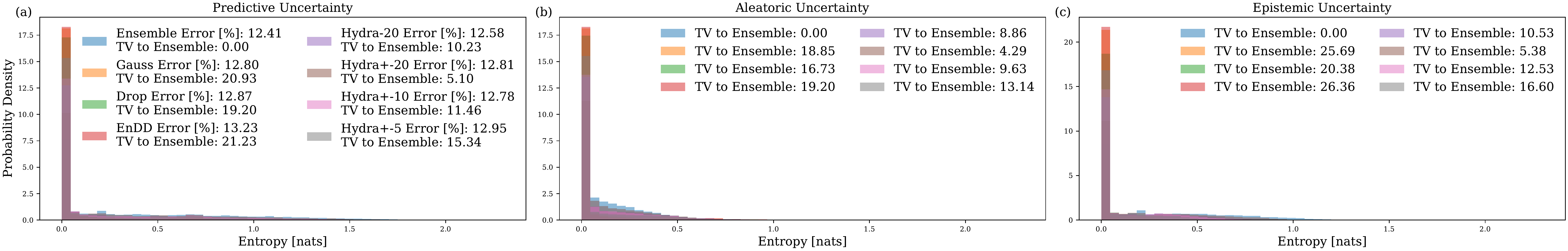}
    \includegraphics[width=1\linewidth]{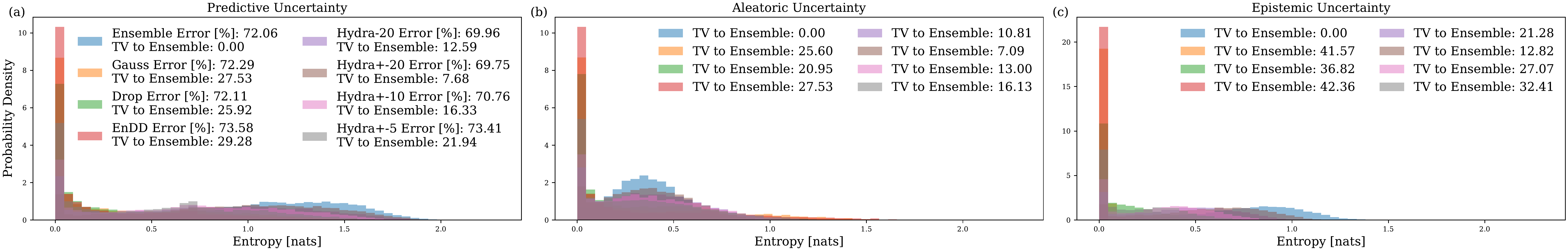}
    \includegraphics[width=1\linewidth]{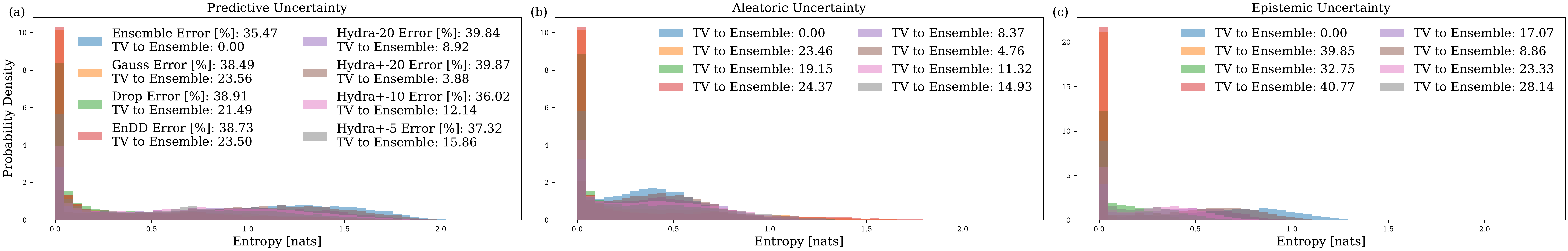}
    \vspace{-2em}
    \caption{Uncertainty decomposition for augmentations applied to the CIFAR-10 test set and the compared methods for one seed/experiment. From the top, in the first row (a-c) is increasing the brightness by 30\%, second row (a-c) is the 45° rotation and the third row (a-c) is 30\% vertical shift. TV denotes total variation and Error denotes the error on the augmented test set.}
    \label{fig:cifar}
\end{figure*}

We adopt the rescaled (0-1) cosine (cos) similarity as the main measure for pushing the weights apart from each other. The rationale behind using the cosine similarity is that it is a proper, differentiable distance metric which is irrespective of magnitude of the weights. By minimising this objective, each head's weights try to move away from the average of the heads' weights at that given level. Thus, the response of each head is induced to be different to all the other ones. Additionally, by comparing the head weights to their abstract mean, instead of a pair-wise comparison, we reduce the compute demand from $\mathcal{O}(M^2)$ to $\mathcal{O}(M)$. On the lowest level, the similarity is implemented between the weights of each separate node or filter, if considering the convolution operation. Related arguments for inducing diversity in NNs were reviewed in~\cite{Gong_2019}. However, to the best of our knowledge, there are no other related works to minimising the similarity between heads in multi-headed KD focused on capturing decomposable uncertainty estimation. 

\subsection{Complete Loss}

Finally, Eqs.~\ref{eq:l1_class} \&~\ref{eq:l1_reg},~\ref{eq:l2_class} \&~\ref{eq:l2_reg},~\ref{eq:l3_class} \&~\ref{eq:l3_reg} and~\ref{eq:l4} can be merged together, with a slight abuse of notation, to give a differentiable optimisation objective $L$: 

\vspace{-2em}

\begin{equation}
    L = (1-\alpha)L_1 + \alpha((1-\beta) L_2 + \beta L_3) + \lambda L_4
    \label{eq:l_complete}
\end{equation}

\vspace{-0.5em}

which is optimised with respect to $\mathbf{\theta}_S$, the teacher responses and the training dataset $D$ via gradient descent in one training session for both classification and regression. The $0 \leq \alpha, \beta\leq 1$ and $\lambda\geq 0$ are hyperparameters dividing the focus between the loss components. $\lambda$ can be kept constant from the start of training or increased linearly from and to a certain iteration. Note that, if $T_{ind},T_{mean} >1$, the respective loss components are being rescaled through $T_{ind}^2$ or $T_{mean}^2$ as discussed in~\cite{hinton2015distilling}.

\section{Experiments}\label{sec:experiments}
In this Section we evaluate the outlined methodology on different experiments. Moreover, we present ablations to the hyperparameter choices and we discuss known limitations that could be targeted in the future work. We varied the dataset and architecture choices to change the complexity of the experiments from toy-data to CIFAR-10 and from feed-forward NNs to ResNet-18 in classification and regression to thoroughly test all the discussed methods.

\paragraph{Hyperparameter Settings} First, we discuss shared hyperparameters across all methods for a fair comparison. Dropout rate was set to 0.5 when used in \textit{Drop} and $\gamma$ for regularisation of KL-divergence in \textit{Gauss} was set to $1/\text{training set size}$ with zero mean, unit variance mean-field prior. No data augmentation was used in training except normalisation. No fine-tuning with or without extra data after training was allowed for any methods. All training had to be performed in one session. For \textit{Hydra}, \textit{Drop} and \textit{Gauss}, $\beta=1.0$, for \textit{Hydra} $\alpha=1.0$ and for \textit{EnDD} $\beta=0$. Further settings for all experiments are described in the Appendix~\ref{sec:appendix}.

\paragraph{Metrics} Second, we discuss the metrics. For classification, we were observing the classification error and expected calibration error (ECE)~\cite{guo2017calibration} with 10 uniformly-spaced bins. For regression, we were measuring the negative log-likelihood (NLL) for a Gaussian output. Specifically, for image-based experiments, we visually and quantitatively compare the decompsoed uncertainty for all test set predictions according to Eq.~\ref{eq:unc_decomposition_class} in normalised histograms and their total variation (TV): $TV(H_a, H_b)=\sum_{i=1}^B |H(i)_b-H(i)_b|$, where $H_a, H_b$ are histograms with $B$ bins. We used $B=50$ bins. From the hardware perspective, we were comparing the number of parameters (\#Params) or required floating-point operations (\#FLOPS).

\subsection{Toy Experiments}

\paragraph{Classification} In the toy classification experiment, seen in the first part of Figure~\ref{fig:toy_example}, we constructed simple feed-forward NNs with 3 hidden layers and ReLU activations, where the last 2 layers served as the heads for \textit{Hydra} and \textit{Hydra+}. As it can be seen, without observing any additional data or fine-tuning, \textit{Hydra+} was able to be at least as uncertain as the teacher ensemble, while having only $228560/618060$, $37\%$ of the parameters and $230860/626060$ of FLOPS when compared to an ensemble when $N, M=20$. As seen in Figure~\ref{fig:toy_example_spiral_appendix}, the method can also near the quality of uncertainty decomposition of the original ensemble.

\paragraph{Regression}  Furthermore, we demonstrate the applicability of the examined method on regression as seen in the second part of Figure~\ref{fig:toy_example}. We constructed a feed-forward NN with 2 hidden layers and ReLU activations where again the last 2 layers served as the heads for \textit{Hydra} and \textit{Hydra+}. In the unobserved regions, \textit{Hydra+} was able to capture the uncertainty of the ensemble, while being able to generalise better than the ensemble, seen in the region where the input $x<-6$. Through $N,M=20$ for \textit{Hydra+} or \textit{Hydra}, the number of parameters or FLOPS were reduced from $106040$ or $109040$ for the ensemble to $55690$ or $56790$ for the student, denoting roughly a 48\% decrease in the required computations and memory consumption for parameters.

\subsection{Image-based Experiments}

For the image-based experiments, we compared their performance under uncertainty on the test data and its augmentations through changes of brightness, rotations or vertical shifts. Additional details, along with experiments, are in the Appendix~\ref{sec:appendix}. All experiments were end-to-end repeated with 3 different random seeds for robustness.

\paragraph{SVHN} The SVHN results are presented in Figure~\ref{fig:svhn} and Table~\ref{tab:svhn}. For SVHN experiments, we adapted the LeNet architecture where the last fully-connected layers served as the heads for \textit{Hydra} or \textit{Hydra+}. Examining results in Table~\ref{tab:svhn}, it can be seen that mainly \textit{Hydra-20} and \textit{Hydra+-20} with $M=20$ heads were able to come close to the performance of the ensemble, seen in error or calibration, while having approximately 13\% of the FLOPS and 56\% of the parameters of the ensemble. Primarily, as seen in Figure~\ref{fig:svhn} \textit{Hydra+-20} was able to significantly reduce the TV to the ensemble in all uncertainty types and in some instances, rows 2 and 3, improve on the error, also improving the overall calibration of the model. However, as seen in the Table~\ref{tab:svhn} or Figure~\ref{fig:svhn}, reducing the number of heads to 10 or 5 has a detrimental effect on the overall performance. From the runtime perspective, including $L_4$ regularisation increased the training cycle by ${\sim}43\%$ without any particular hardware optimisations if compared to \textit{Hydra}.

\paragraph{CIFAR-10}
The CIFAR-10 results are presented in Figure~\ref{fig:cifar} and Table~\ref{tab:cifar_10}. For CIFAR-10 experiments we adapted the ResNet-18 where the last 2 blocks served as the heads for \textit{Hydra} or \textit{Hydra+}. As it can be seen from the Table~\ref{tab:cifar_10}, given the complexity of the task, no method was able to achieve lower error than the ensemble. However \textit{Hydra+} with $M=20$ was able to be better calibrated than the ensemble with $37\%$ of the FLOPS and $34\%$ of the parameters. Interestingly, as the number of heads was decreased the error did not significantly deteriorate, but due to the reduced capacity, the representation power was smaller, primarily seen in Figure~\ref{fig:cifar}. In Figure~\ref{fig:cifar} it can be seen that the examined loss in Eq.~\ref{eq:l_complete} and regularisation were able to significantly decrease TV between the teacher and the student for predictive, aleatoric and epistemic uncertainty, especially when compared \textit{Hydra-20} to \textit{Hydra+-20}. In a more complex model, compared to the SVHN experiment, including $L_4$ regularisation prolonged the training by ${\sim}6\%$ when compared to \textit{Hydra}, given that the most of the parameters are in convolutions instead of fully-connected layers as in the SVHN experiment.

\begin{table}[t!]
  \centering
  \scalebox{0.77}{
  \begin{tabular}{@{}lcccc@{}}
    \toprule
    \textbf{Method} & \textbf{Error} [\%] & \textbf{ECE} [\%] & \textbf{\#FLOPS} [M] &  \textbf{\#Params} [M]  \\
    \midrule
   Ensemble-20 & 7.20$\pm$0.05 & 3.81$\pm$0.11& 117.75& 3.04\\
    Gauss  & 8.53$\pm$0.21& 6.08$\pm$0.13& 13.48 &0.13\\
    Drop &  8.70$\pm$0.14 &5.17$\pm$0.08 & 15.07&0.12\\
    EnDD &  8.91$\pm$0.08 & 6.45$\pm$0.11& 13.48 &0.12\\
    Hydra-20 &7.49$\pm$0.06 &3.10$\pm$0.10&15.07 & 1.71\\ \hline
    Hydra+-20  &7.56$\pm$0.06 &3.08$\pm$0.03 &15.07 & 1.71\\
    Hydra+-10  &7.59$\pm$0.04 & 3.14$\pm$0.01 & 14.23 &0.88\\
    Hydra+-5  & 7.67$\pm$0.08 & 3.69$\pm$0.08 & 13.81 &0.46\\
    \bottomrule
  \end{tabular}}
  \vspace{-0.5em}
  \caption{Comparison on the test dataset of SVHN. The number after method name denotes $N$ or $M$.}
  \label{tab:svhn}
\end{table}
\begin{table}[t!]
    \vspace{-0.5em}
  \centering
  \scalebox{0.77}{
  \begin{tabular}{@{}lcccc@{}}
    \toprule
    \textbf{Method} & \textbf{Error} [\%] & \textbf{ECE} [\%] & \textbf{\#FLOPS} [G] &  \textbf{\#Params} [M] \\
    \midrule
   Ensemble-20 & 10.92$\pm$0.10 & 4.62$\pm$0.20  & 2.81 & 55.95\\
    Gauss  & 11.70$\pm$0.10 &9.53$\pm$0.09 &0.80 & 4.55\\
    Drop & 11.61$\pm$0.19 &8.84$\pm$0.14 & 0.80 & 4.55\\
    EnDD & 11.80$\pm$0.29  & 9.70$\pm$0.34& 0.80 & 4.55 \\
    Hydra-20  & 11.44$\pm$0.28 & 4.83$\pm$0.14&1.04& 19.23\\ \hline
    Hydra+-20  & 11.54$\pm$0.16 & 3.74$\pm$0.10 &1.04& 19.23\\
    Hydra+-10  & 11.37$\pm$0.05 & 4.98$\pm$0.07& 0.92&11.51\\
    Hydra+-5  & 11.29$\pm$0.07& 6.47$\pm$0.11& 0.85 &7.64\\
    \bottomrule
  \end{tabular}}
  \vspace{-0.5em}
  \caption{Comparison on the test dataset of CIFAR-10. The number after method name denotes $N$ or $M$.}
  \label{tab:cifar_10}
\end{table}

\subsection{Ablations}\label{sec:experiments_ablations}

\begin{figure}[t]
    \centering
    \includegraphics[width=1.0\linewidth]{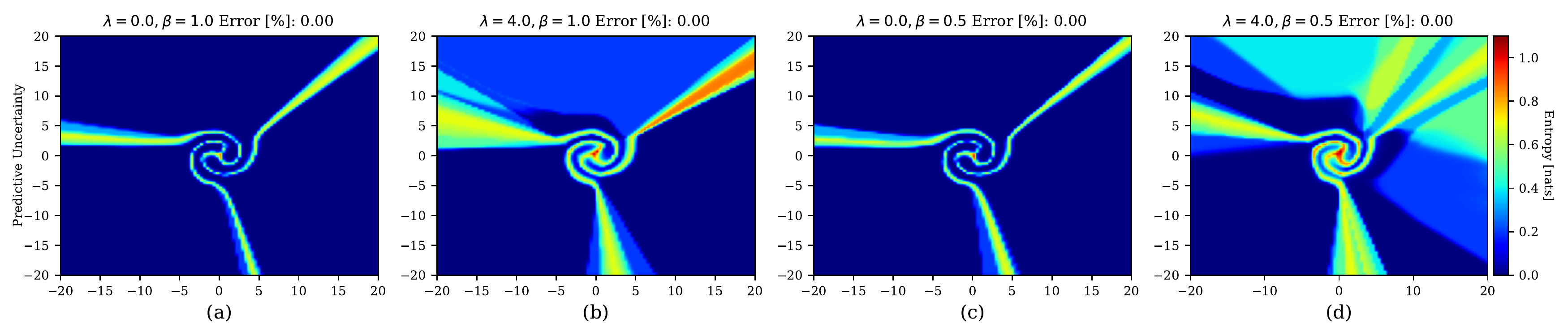}
    \vspace{-2em}
    \caption{From the left to right, (a) disabling $L_4$, $\beta=1.0$, (b) enabling $L_4$, $\beta=1.0$, (c) disabling $L_4$, $\beta=0.5$, (d) enabling $L_4$, $\beta=0.5$ for the toy classification problem. }
    \label{fig:toy_example_spiral_ablation}
    \vspace{-1em}
\end{figure}

\begin{figure}[t]
    \centering
    \includegraphics[width=1.0\linewidth]{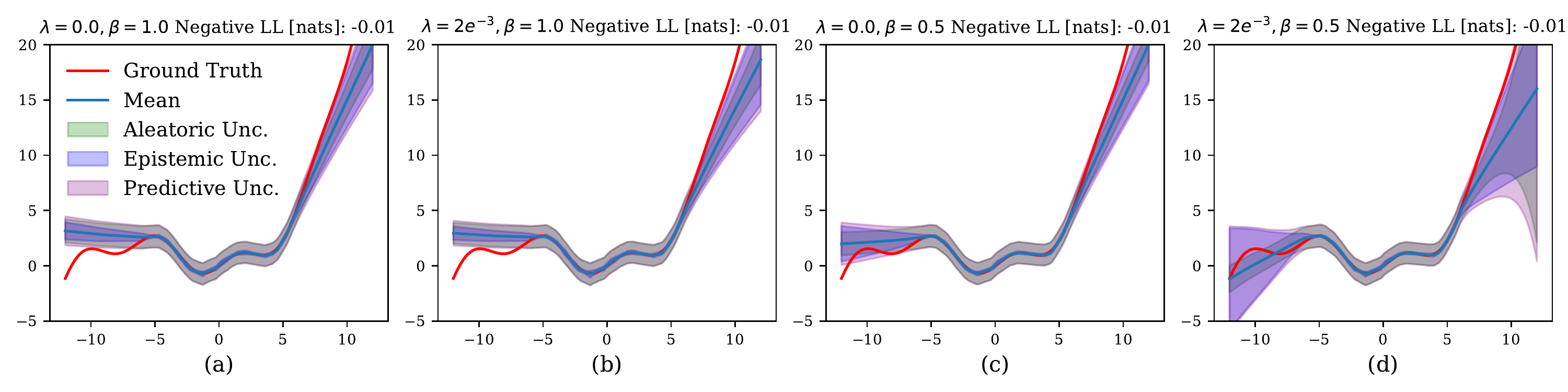}
    \vspace{-2em}
    \caption{From the left to right, (a) disabling $L_4$, $\beta=1.0$, (b) enabling $L_4$, $\beta=1.0$, (c) enabling $L_4$, $\beta=0.5$, (d) enabling $L_4$, $\beta=0.5$ for the toy regression problem.}
    \label{fig:toy_example_regress_ablation}
\end{figure}

\paragraph{Changing $\lambda, \beta$}
Additionally, we wanted to demonstrate the effects of changing $\lambda$ and $\beta$ hyperparameters for the student models. For a clear visualisation of the responses, the changes are illustrated on the toy problems as seen in Figures~\ref{fig:toy_example_spiral_ablation} \&~\ref{fig:toy_example_regress_ablation}.  In general, we observe that the main portion of the observed epistemic uncertainty is caused by including the $L_4$ term in the loss function through $\lambda$, as seen in subplots (b) in both Figures~\ref{fig:toy_example_spiral_ablation} \&~\ref{fig:toy_example_regress_ablation}. Conversely, $\beta$ being less than 1 and thus including $L_2$ in the optimisation, guides the overall fit of the model and rectifies its predictive uncertainty, most notably the aleatoric component, which could otherwise result in uncalibrated predictions, as seen if comparing Figures~\ref{fig:toy_example_spiral_ablation} \&~\ref{fig:toy_example_regress_ablation} (a,c)  in the regions close to the origin, where training data was actually observed. Interestingly, we observed that $L_3$ and $L_4$ are closely related and empirically we were unable to obtain as good results as seen in Figures~\ref{fig:toy_example_spiral_ablation} \&~\ref{fig:toy_example_regress_ablation} (d) without enabling both $L_3$ and $L_4$ during training. 


\begin{figure}[t]
    \centering
    \includegraphics[width=1.0\linewidth]{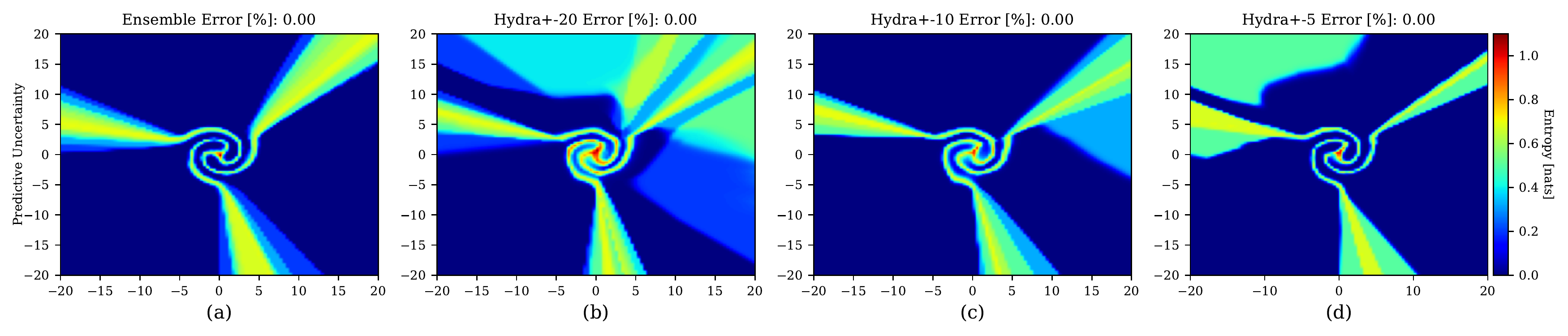}
    \vspace{-2em}
    \caption{From the left to right, (a) Ensemble, (b) $M=20$, (c) $M=10$, (d) $M=5$ for the toy classification problem. }
    \label{fig:toy_example_spiral_heads}
    \vspace{-2.em}
\end{figure}

\begin{figure}[t]
    \centering
    \includegraphics[width=1.0\linewidth]{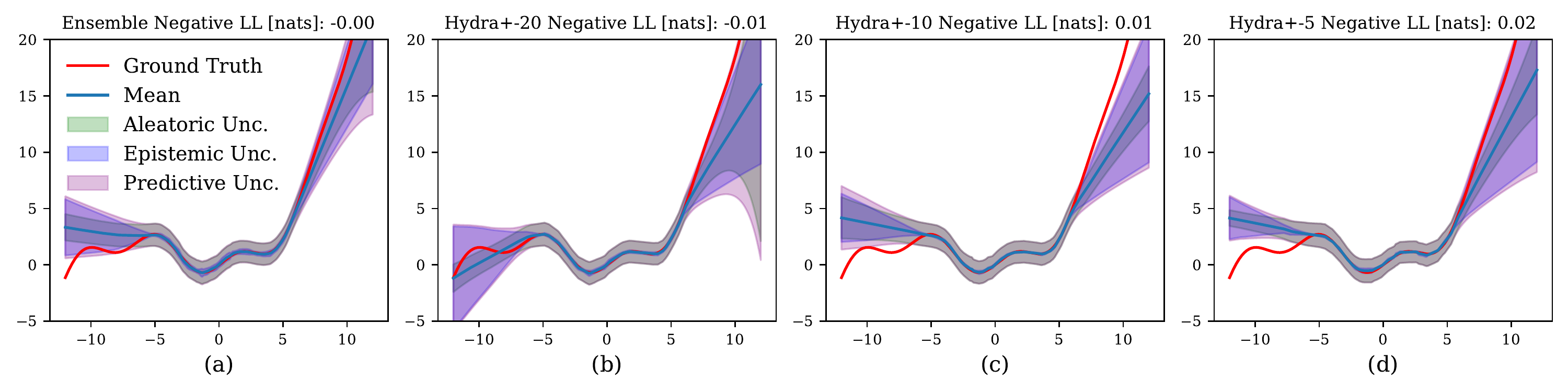}
    \vspace{-2em}
    \caption{From the left to right, (a) Ensemble, (b) $M=20$, (c) $M=10$, (d) $M=5$ for the toy regression problem.}
    \label{fig:toy_example_regress_heads}
\end{figure}

\paragraph{Changing $M$}
Next, we discuss changing the number of heads in \textit{Hydra+} through decreasing $M\leq N$. For the toy datasets, this is visualised in Figures~\ref{fig:toy_example_spiral_heads} \&~\ref{fig:toy_example_regress_heads} and for CIFAR-10 and SVHN experiments in Tables~\ref{tab:svhn} \&~\ref{tab:cifar_10} or Figures~\ref{fig:svhn} \&~\ref{fig:cifar}. Decreasing $M$ corresponds to decreasing the representation ability of the student, through practically decreasing its number of parameters and pushing a head to learn from multiple ensemble members. As a result, the smaller, less representative students' uncertainty representation power deteriorates. Nevertheless, their accuracy, does not necessarily need to follow the same trend, where a head learns to generalise better if focusing on more than one teacher. This is best seen in Table~\ref{tab:cifar_10} for the CIFAR-10 experiment. Additionally, having to run fewer heads on hardware reduces the required computation and memory storage as seen in Tables~\ref{tab:svhn} \&~\ref{tab:cifar_10} when comparing FLOPS or the number of parameters. 


\subsection{Discussion, Limitations \& Future Work}\label{sec:experiments_cifar_discussion}

\paragraph{Discussion} 

\citet{mehrtens2022improving}, considering~\cite{benjamin2018measuring} came to the conclusion that parameter distance is no good measure for functional differences between ensemble members. Nevertheless,~\citet{mehrtens2022improving} suggest that in certain instances~\cite{wenzel2020hyperparameter}, where parameters are not easily separable from each other, increasing orthogonality in parameter space could induce functional diversity. We believe that this particular setting, with KD and a shared core and intra-dependent training of the heads is another such case. We did try the distance mentioned~\cite{mehrtens2022improving}, adapted to multiple heads instead of ensemble members, or different cosine-based distance measures~\cite{sohangir2017improved}, but empirically they did outperform the proposed measure quantitatively or qualitatively. We used $N=20$ in the ensemble, such that it had strong representation power. We empirically observed that if $N$ is small e.g. $5$, irrespective of the hyperparameters, the students are unable to capture barely any uncertainty representation. Moreover, we noticed that if the $N$ ensemble members are randomly shuffled for Eqs.~\ref{eq:l3_class} \&~\ref{eq:l3_reg}, the heads are only able to retain general representation without any individuality, avoiding the capture of any uncertainty estimation of the teacher. We also empirically observed that \textit{Hydra+} performs better when $L_4$ was not applied to batch normalisation weights. We empirically observed that it is beneficial to increase $\lambda$ as $M$ or $N$ decrease. In certain instances, e.g. the CIFAR-10 experiment, it was better to apply $L_4$ later in the training and increase its influence as the training progresses. Similarly to~\citet{tran2020hydra}, we empirically observed that increasing the individual head's size, the depth or width, improves their representation power, but it also increases the overall number of required parameters and FLOPS.

In observing the performance of \textit{Gauss}, \textit{Drop} or \textit{EnDD} we see a trend that if the prior distribution is potentially misspecified the student network is unable to capture the uncertainty estimation of the teacher. This gives preference, to rather simpler, however more compute expensive, from the parameter of FLOPS standpoint, distribution-free methods such as \textit{Hydra} which does not require any particular assumptions, but only hyperparameter tuning for the provided loss function. 

\paragraph{Limitations}

In our experimentation we observed several corner cases. To begin with, at the moment it is necessary to manually define $M$. The architecture of the student was manually selected and it had to be hand-tuned on the validation dataset, along with other hyperparameters. Moreover, the examined regularisation focuses primarily on improving the estimation of the epistemic uncertainty, without the ability to control the aleatoric uncertainty other than relying on the performance of teacher. Last but not least, other loss functions could be examined which would focus on capturing the uncertainty estimation more prominently in comparison to KL divergence.

\paragraph{Future Work}

As seen for more challenging tasks, such as SVHN or CIFAR-10 classification, it is difficult to match the performance of the ensemble. We believe that to provide further adaptation through diversity in both the function space~\cite{mehrtens2022improving} and parameter space, the student architecture could be optimised in addition to the parameters. Our current efforts are in applying neural architecture search~\cite{chen2020drnas} to find the architecture of the required $M$ heads and core automatically. We are also experimenting in automatising the search for $M$, such that we can further reduce the FLOPS count or the number of parameters.

\section{Conclusion}\label{sec:conclusion}

In this work-in-progress we examined a regularisation for knowledge distillation, to capture both the generalisation performance as well as uncertainty estimation of the teacher ensemble without any fine-tuning or extra data. We demonstrated the explored methodology on toy and real-world data and different architectures to show its versatility. In comparison to the underlying ensembles, the discussed regularisation was able to approach near or improve upon its quality of calibration and uncertainty estimation. In the future work, we aim to improve the student's performance through automatic adaptation of the student architecture to the task and the teacher network.
\section*{Acknowledgements}
Martin Ferianc was sponsored through a scholarship from the Institute of Communications and Connected Systems at UCL and through the PhD Enrichment scheme at The Alan Turing Institute. Lastly, we thank DFUQ'22 reviewers for feedback and encouragement. 

\bibliography{bib}
\bibliographystyle{src/icml2022}

\appendix
\section{Appendix}\label{sec:appendix}

\subsection{Further Experimental Settings}

All experiments were performed with respect to 200 epochs, batch size 256, Adam optimiser with default $\beta_1, \beta_2$ hyperparameters, with cosine decreasing learning rate schedule and gradient clipping coefficient set to 5.0. 10\% of any training set was set for validation to hand-tune the hyperparameters. PyTorch 1.11, CUDA 11.1 and Nvidia GeForce RTX 2080 SUPER GPU were used for implementation. The default PyTorch initialisation was used for the weights. Note that, we reimplemented the weight decay to be computed explicitly and then added to the loss and we did not use the default weight decay option for optimisers in PyTorch. The ensemble was always trained simply with respect to a cross-entropy loss or Gaussian negative log-likelihood. The $M$ in context of \textit{Gauss}, \textit{Drop} or \textit{EnDD} meant the number of Monte Carlo samples for training or evaluation. We have combined the predictions of the student or the teacher with respect to all $N$ or $M$ samples and averaged, after the softmax activation for classification and for regression, we combined the variance through Eq.~\ref{eq:unc_decomposition_reg}. 

For \textit{Hydra} or \textit{Hydra+} we initialised the architecture already with $M$ heads to observe the performance after a singular training session. We wanted to benchmark the methods for the simplicity of implementation of a single training session and to avoid interference from multiple steps and choices of hyperparameters that would be needed if the original \textit{training with multi-head growth} was to be considered. This is also reflected in the choices of other hyperparemeters, where we did not want to disadvantage any other method, such that we can provide a fair comparison and outlook.

The image corruption experiments consisted of [20, 30, 50]\% increases in brightness intensity, [15, 45, 75] degree rotations, [20, 30, 50]\% vertical shifts for both datasets. We encourage the reader to see our code, how to generate additional results for corruptions not shown in the paper. 

\subsection{Toy Classification}

\begin{figure}[t]
    \centering
    \includegraphics[width=0.95\linewidth]{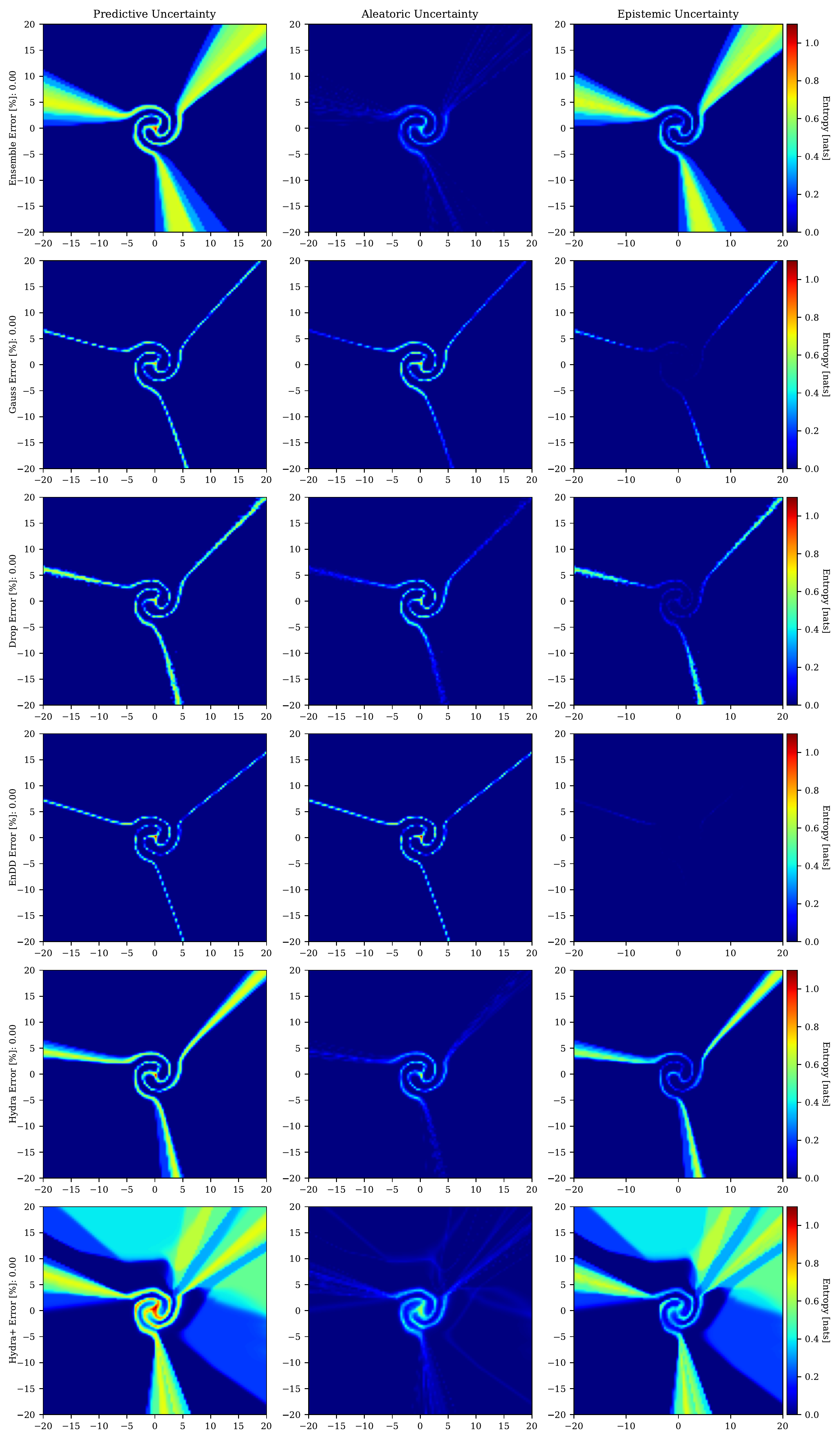}
    \vspace{-1.em}
    \caption{Decomposition of the uncertainty into aleatoric and epistemic parts for the toy classification spiral dataset and the compared methods from Figure~\ref{fig:toy_example}.}
    \label{fig:toy_example_spiral_appendix}
\end{figure}

The toy classification experiment was performed with respect to a feed-forward network with [2, 100, 100, 100, 100, 3] input, hidden nodes and output, ReLU activations, initial learning rate 0.01, weight decay applied to the teacher training was set to $1e^{-4}$. The training, validation and test sets contained 240, 30, 30 points split equally across 3 classes sampled from 3 spirals originating at coordinates 0,0. For the KD of the student the weight decay was set to $1e^{-8}$, $\alpha=0.9$, $T_{ind} = 3.0$, $T_{mean} = 1.0$ and $\beta$ for \textit{Hydra+} was set to 0.5. For 20, 10 and 5 heads of \textit{Hydra+}, the $\lambda$ was set to $4.0, 7.0$ and $9.0$ respectively.

\begin{figure}[t]
    \centering
    \includegraphics[width=0.95\linewidth]{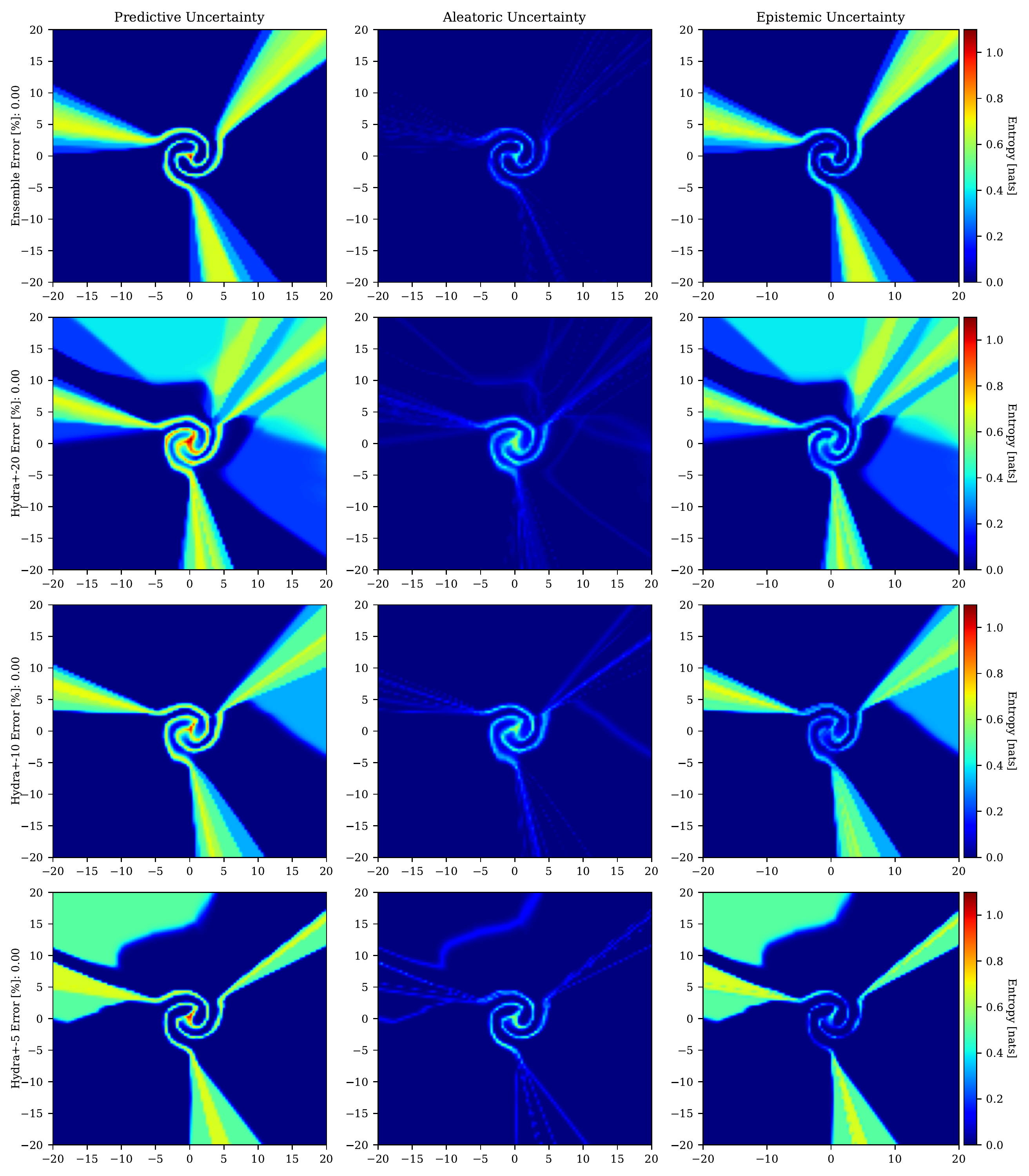}
    \vspace{-1.em}
    \caption{Decomposition of the uncertainty into aleatoric and epistemic parts for the toy classification spiral dataset and varying number of heads from Figure~\ref{fig:toy_example_spiral_heads}.}
    \label{fig:toy_example_spiral_heads_appendix}
\end{figure}

\begin{figure}[t]
    \centering
    \includegraphics[width=0.95\linewidth]{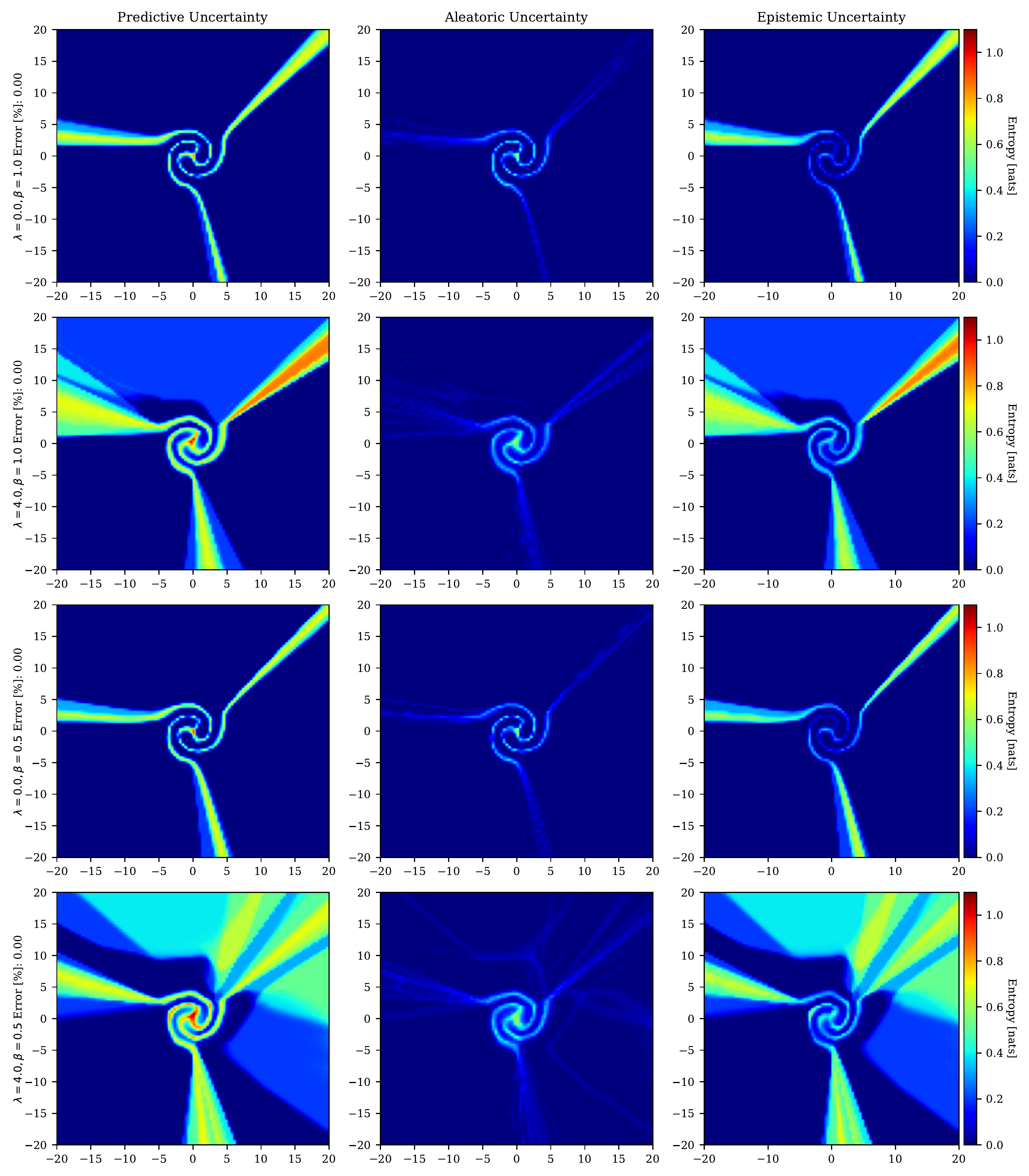}
    \vspace{-1.em}
    \caption{Decomposition of the uncertainty into aleatoric and epistemic parts for the toy classification spiral dataset and ablations from Figure~\ref{fig:toy_example_spiral_ablation}.}
    \label{fig:toy_example_spiral_ablation_appendix}
\end{figure}

Figure~\ref{fig:toy_example_spiral_appendix} shows further decomposition of uncertainty for the compared methods according to Eq.~\ref{eq:unc_decomposition_class}. Figure~\ref{fig:toy_example_spiral_heads_appendix} shows further decomposition of uncertainty for the varying number of heads. Figure~\ref{fig:toy_example_spiral_ablation_appendix} shows further decomposition of uncertainty for the varying hyperparameters as discussed in the ablations in Section~\ref{sec:experiments_ablations}.


\subsection{Toy Regression}

The toy regression experiment was performed with respect to a feed-forward network with [2, 50, 50, 50, 2] input, hidden nodes and output, ReLU activations, initial learning rate 0.05, weight decay applied to the teacher training was set to $1e^{-5}$. The training, validation and test sets contained 240, 30, 30 points sampled from $f(x) =sin(x)-0.1x + 0.1x^{2} + 0.01x^{3}$ from range $-6$ to $6$ with additive Gaussian noise $\mathcal{N}(0,1)$ added to the training points. For the KD of the student the weight decay was set to $1e^{-8}$, $\alpha=0.9$ and $\beta$ for \textit{Hydra+} was set to 0.5. For 20, 10 and 5 heads of \textit{Hydra+}, the $\lambda$ were linearly increased from $0.0$, beginning on the 50\textsuperscript{th} to $2e^{-3}, 0.02$ or $0.6$ culminating at the 150\textsuperscript{th} epoch respectively.

\subsection{SVHN}

The SVHN classification experiment was performed with respect to a LeNet-like architecture with 2 convolutions, followed by ReLU and maxpooling with 2 fully-connected layers with [3, 32, 64, 1024, 128, 10] input, hidden and output channels for the teacher. The student had a similar architecture, however, with [3, 128, 32, 512, 128, 128, 10] input, hidden and output split. The learning rate was initialised to 0.001, weight decay applied to the teacher training was set to $1e^{-4}$. The training, validation and test sets contained 65932, 7325 and 26032 samples split across 10 classes. For the KD of the student the weight decay was set to $1e^{-8}$, $\alpha=0.95$, $T_{ind} = 5.0$, $T_{mean} = 1.0$ and $\beta$ for \textit{Hydra+} was set to 0.9. For 20, 10 and 5 heads of \textit{Hydra+}, the $\lambda$ were set to $5e^{-4}, 1e^{-3}, 5e^{-3}$ respectively.

\begin{figure*}[b]
    \centering
    \includegraphics[width=1\linewidth]{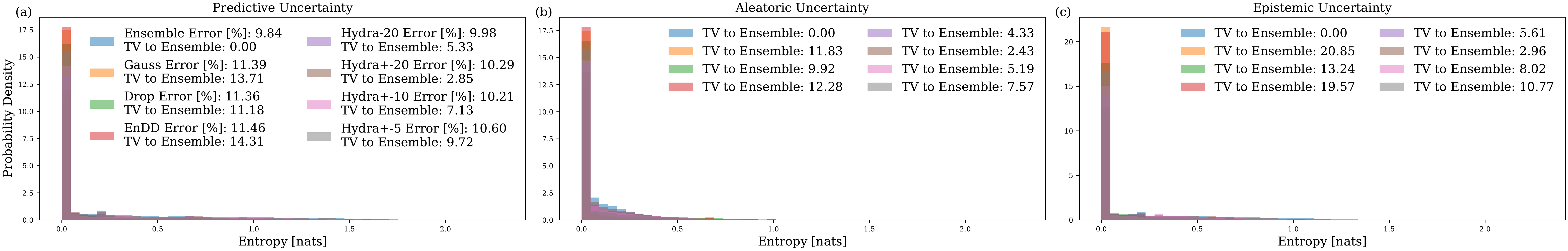}
    \includegraphics[width=1\linewidth]{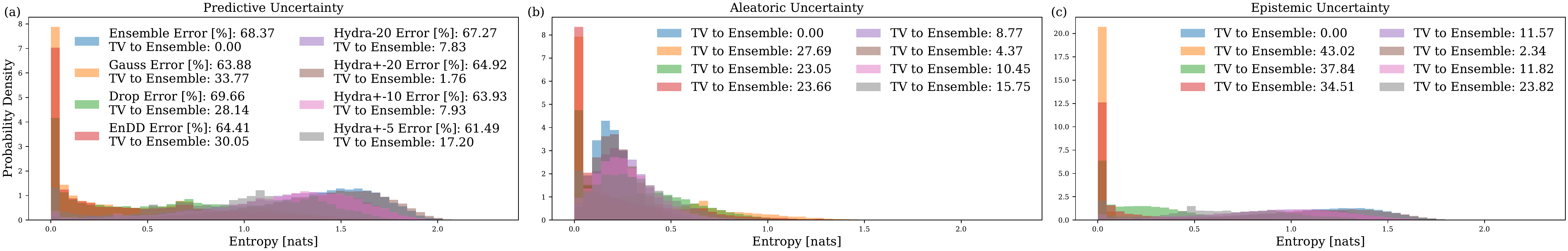}
    \includegraphics[width=1\linewidth]{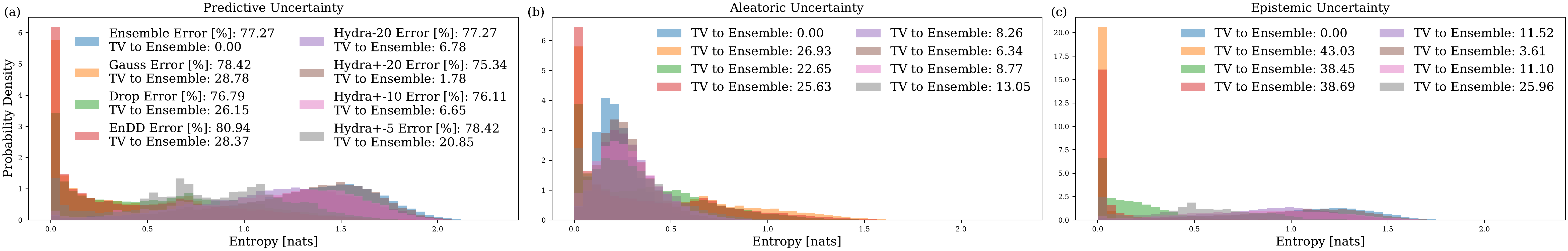}
    \vspace{-2em}
    \caption{Uncertainty decomposition for augmentations applied to the SVHN test set and the compared methods for one seed/experiment. From the top, in the first row (a-c) is increasing the brightness by 50\%, second row (a-c) is the 30° rotation and the third row (a-c) is 30\% vertical shift. TV denotes total variation and Error denotes the error on the augmented test set.}
    \label{fig:svhn_appendix}
\end{figure*}

Figure~\ref{fig:svhn_appendix} shows further decomposition of uncertainty for the compared methods according to Eq.~\ref{eq:unc_decomposition_class} for more severe augmentations.

\subsection{CIFAR-10}

The CIFAR-10 classification experiment was performed with respect to a ResNet-18 architecture with 8 residual blocks and output channel sizes [32, 64, 128, 256]. The student had a similar architecture, however, with [96, 128, 256, 128] output channel sizes. The learning rate was initialised to 0.01, weight decay applied to the teacher training was set to $1e^{-4}$. The training, validation and test sets contained 45000, 5000 and 10000 samples split across 10 classes. For the KD of the student the weight decay was set to $1e^{-8}$, $\alpha=0.95$, $T_{ind} = 8.0$, $T_{mean} = 1.0$ and $\beta$ for \textit{Hydra+} was set to 0.5. For 20, 10 and 5 heads of \textit{Hydra+}, the $\lambda$ were linearly increased from $0.0$, beginning on the 20\textsuperscript{th} to $2e^{-3}, 0.05$ or $0.1$ culminating at the 150\textsuperscript{th} epoch respectively.

\begin{figure*}[t]
    \centering
    \includegraphics[width=1\linewidth]{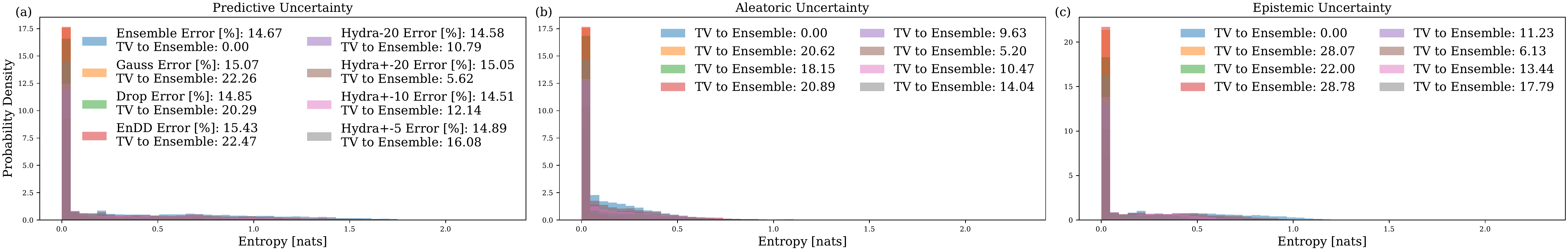}
    \includegraphics[width=1\linewidth]{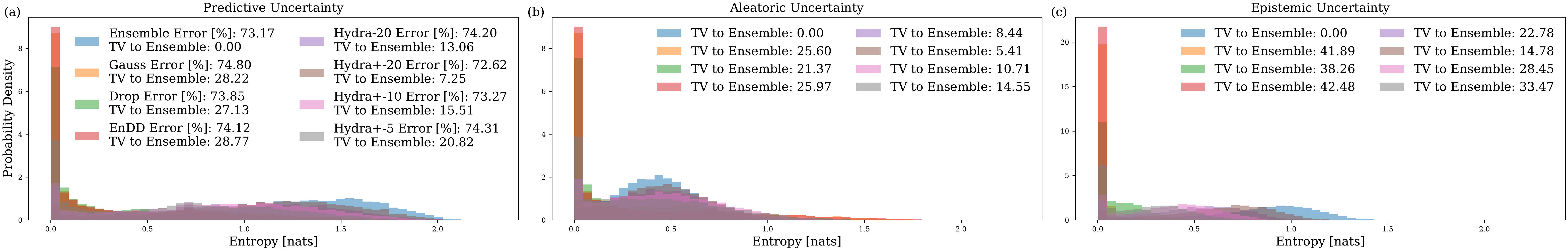}
    \includegraphics[width=1\linewidth]{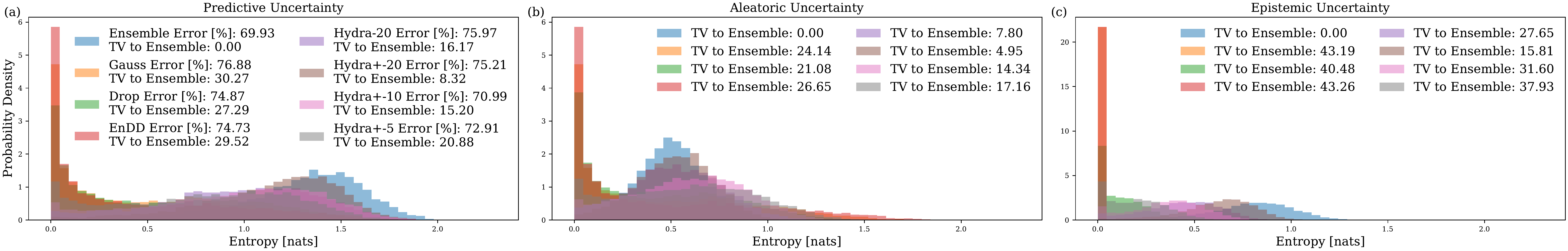}
    \vspace{-2em}
    \caption{Uncertainty decomposition for augmentations applied to the CIFAR-10 test set and the compared methods for one seed/experiment. From the top, in the first row (a-c) is increasing the brightness by 50\%, second row (a-c) is the 75° rotation and the third row (a-c) is 50\% vertical shift. TV denotes total variation and Error denotes the error on the augmented test set.}
    \label{fig:cifar_appendix}
\end{figure*}

Figure~\ref{fig:cifar_appendix} shows further decomposition of uncertainty for the compared methods according to Eq.~\ref{eq:unc_decomposition_class} for more severe augmentations.

\end{document}